\documentclass[a4paper,12pt]{article}

\usepackage[utf8]{inputenc}
\usepackage[T1]{fontenc}
\usepackage{amsmath}
\usepackage{amssymb}
\usepackage{amsthm}
\usepackage{amsfonts}
\usepackage{graphicx}
\usepackage{tikz}
\usepackage{xspace}
\usepackage{hyperref}
\usepackage{xcolor}
\usepackage{algorithm}
\usepackage{algorithmic}
\usepackage{mathtools}
\usepackage[autolanguage,np,boldmath]{numprint}
\usepackage{booktabs}

\usetikzlibrary{calc,3d}

\newcommand{\naturals}{\mathbb{N}}

\newcommand{\field}{\mathbb{F}}
\newcommand{\binarySet}{\{0,1\}}

\newcommand{\range}[2]{[#1..#2]}

\renewcommand{\vec}[1]{\boldsymbol{#1}}

\begin{document}

\title{Linear representation of categorical values\footnote{An
    extended two-page abstract of this work will appear in \emph{2021
      Genetic and Evolutionary Computation Conference Companion (GECCO
      '21 Companion)}. \url{https://doi.org/10.1145/3449726.3459513}}}
\author{Arnaud Berny}
\maketitle

\begin{abstract}
  We propose a binary representation of categorical values using a
  linear map. This linear representation preserves the neighborhood
  structure of categorical values. In the context of evolutionary
  algorithms, it means that every categorical value can be reached in
  a single mutation. The linear representation is embedded into
  standard metaheuristics, applied to the problem of Sudoku puzzles,
  and compared to the more traditional direct binary encoding. It
  shows promising results in fixed-budget experiments and empirical
  cumulative distribution functions with high dimension instances, and
  also in fixed-target experiments with small dimension instances.
\end{abstract}


\section{Introduction}

Representation is an important topic for evolutionary algorithms
\cite{rothlauf2006representations} and other metaheuristics,
especially when applied to combinatorial optimization. It directly
influences the range of problems which can be adressed by
metaheuristics and the quality of their solutions. Many evolutionary
algorithms have been designed with binary domains in mind. Although
most of them can be adapted more or less easily to other domains, it
is still desirable to be able to represent values from non binary
domains in binary domains so as to leverage theoretical and practical
knowledge of evolutionary algorithms in binary domains along with
their implementations. In this paper, we are concerned with the binary
representation of categorical values.

Often, categorical values are represented by means of a direct binary
encoding. As an example, let us address the problem of representing
the four nucleobases $A$, $T$, $C$, and $G$ found in DNA. Using 2
bits, we can arbitrarily decide that $A = 00$, $T = 01$, $C = 10$, and
$G = 11$. In the neighborhood system defined by 1-bit flips, it
appears that each nucleobasis has 2 neighbors. Hence it is not
possible to go from $A$ to $G$ in a single bit flip. The so called
unary representation is also able to represent categorical values. The
idea is to assign categorical values to bit strings based on their
Hamming weights. In the case of DNA, we can arbitrarily decide that
$A$, $T$, $C$, and $G$ are represented by 3-bit strings of Hamming
weights 0, 1, 2, and 3 respectively. Consequently, $A$ is represented
by only one bit string (000) whereas $C$ is represented by 3 bit
strings (011, 101, and 110). Just as for direct encoding, the
neighborhood system of 1-bit flips is not complete in the sense that
$A$ and $G$ have only one neighbor as compared to $T$ and $C$ which
have two neighbors. Moreover, going from $A$ to $G$ requires 3 bit
flips.

Those representations are inappropriate because the resulting
neighborhood systems among categorical values are not complete. In a
set of categorical values, every element is the neighbor of every
other element. In other words, categorical values are the vertices of
a complete graph. In this paper, we propose a binary representation of
categorical values which is based on a linear map and which satisfies
this requirement.

The paper is organised as follows. In Sect.~\ref{sec:representation}
we define binary representations for categorical values. In
Sect.~\ref{sec:line-repr} we propose a linear representation for
categorical values. In Sect.~\ref{sec:experiments} we apply the linear
representation to Sudoku puzzles seen as optimization problems.
Sec.~\ref{sec:conclusion} concludes the paper.

\section{Representation}
\label{sec:representation}

Let $V = \{v_1, v_2, \ldots, v_N\}$ be a set of $N$ categorical
values, where $N\in\naturals$ and $N\ge 2$. We want to represent those
values in a binary domain $\binarySet^n$ of dimension $n\in\naturals$.
A binary representation of $V$ is a surjective map
$\phi : \binarySet^n \rightarrow V$, that is, for all $v\in V$, there
exists $\vec{x}\in\binarySet^n$ such that $\phi(\vec{x}) = v$. The
binary vector $\vec{x}$ is called a representative of $v$ which might
have more than one representative. Such binary representations can be
used, for example, to apply metaheuristics designed for binary spaces
to the optimization of functions defined on categorical values.

Let $(\vec{e}_1, \vec{e}_2, \ldots, \vec{e}_n)$ be the canonical basis
of $\binarySet^n$. For example, in $\binarySet^3$,
$\vec{e}_1 = (1, 0, 0)^t$, where $t$ denotes transpose (we use column
vectors). For all $\vec{x}\in\binarySet^n$, let $B(\vec{x}, 1)$ be the
Hamming ball of radius 1 centered at $\vec{x}$, that is
$B(\vec{x}, 1) = \{\vec{x}\} \cup \{\vec{x}+\vec{e}_i \mid
i\in\range{1}{n}\}$. Throughout this paper, we identify the set
$\binarySet$ as the finite field $\field_2$. Thus, addition on
$\binarySet$ or $\binarySet^n$ must be understood modulo 2 and is
equivalent to the exclusive-or operator. For example, if
$\vec{x} = (1, 1, 1)^t$ then $\vec{x}+\vec{e}_1 = (0, 1, 1)^t$.

With mutation based metaheuristics or local search in mind, we would
like to be able to reach any categorical value in a single bit
mutation. We say that $\phi$ is locally bijective if, for all
$\vec{x}\in\binarySet^n$, its restriction
$\phi : B(\vec{x}, 1) \rightarrow V$ is bijective. In this case,
necessarily, $n + 1 = N$. We can restate this property in the language
of graph theory by saying that the hypercube $\binarySet^n$ is a
covering graph of the complete graph $K_N$ and $\phi$ a covering map
from $\binarySet^n$ to $K_N$. The map $\phi$ is also an $N$-coloring
of $\binarySet^n$.

\section{Linear representation}
\label{sec:line-repr}

We propose a linear representation which is locally bijective. We
suppose for now that $N = 2^k$, where $k\in\naturals$. The categorical
values are first identified with $k$-bit binary vectors in an
arbitrary manner. We are looking for a surjective linear
representation, that is a $k\times n$ binary matrix of rank $k$. Let
$\vec{x}\in\binarySet^n$ be the current search point and
$\vec{y} = \vec{A}\vec{x}\in\binarySet^k$ its corresponding
categorical value. The neighbors of $\vec{y}$ are
$\vec{A}(\vec{x}+\vec{e}_i) = \vec{A}\vec{x} + \vec{A}\vec{e}_i =
\vec{y} + \vec{A}\vec{e}_i$, where $i\in\range{1}{n}$. Let
$\vec{y'}\in\binarySet^k$ be any categorical value but $\vec{y}$.
Then,
$\vec{A}(\vec{x}+\vec{e}_i) = \vec{y'} \Leftrightarrow
\vec{A}\vec{e}_i = \vec{y} + \vec{y'}$. The last equation has a unique
solution if and only if the set
$\{ \vec{A}\vec{e}_i \mid i\in\range{1}{n} \}$ is the set
$\binarySet^k \setminus \{\vec{0}\}$ and $n = N - 1 = 2^k - 1$, which
means that the columns of $\vec{A}$ are made of all the vectors of
$\binarySet^k$ but $\vec{0}$.

We observe that $\vec{A}$ is precisely the parity-check matrix of the
binary Hamming code
\cite{hamming-1950,pless89:_introd_theor_error_correc_codes}. It is
remarkable that a requirement in the context of local search leads to
a well known object of coding theory. We want to point out that we use
$\vec{A}$ differently, though. The Hamming code is defined as the set
of vectors $\vec{x}$ of $\binarySet^n$ such that
$\vec{A}\vec{x} = \vec{0}$, that is, as the kernel (null space) of
$\vec{A}$. We use $\vec{A}$ more as a generator matrix. The matrix
$\vec{A}$ is also the generator matrix of the simplex code which is
the dual code of the Hamming code. It is defined as the set
$\{\vec{y}^t \vec{A} \mid \vec{y}\in\binarySet^k\}$, where we have
used left-multiplication as is traditional in coding theory
litterature. Again, we use $\vec{A}$ differently.

\begin{figure}
  \centering
  \begin{tikzpicture}[scale=5]
    \draw[dashed, right,->] (0,0) -- (xyz cs:x=1.2) node {$x_1$};
    \draw[dashed, above,->] (0,0) -- (xyz cs:y=1.2) node {$x_2$};
    \draw[dashed, below,->] (0,0) -- (xyz cs:z=1.2) node {$x_3$};
    \coordinate (C000) at (0, 0, 0);
    \coordinate (C001) at (0, 0, 1);
    \coordinate (C011) at (0, 1, 1);
    \coordinate (C010) at (0, 1, 0);
    \coordinate (C110) at (1, 1, 0);
    \coordinate (C111) at (1, 1, 1);
    \coordinate (C101) at (1, 0, 1);
    \coordinate (C100) at (1, 0, 0);
    \draw (C110) -- (C111) -- (C101) -- (C100) -- cycle;
    \draw (C010) -- (C110) -- (C111) -- (C011) -- cycle;
    \draw (C111) -- (C011) -- (C001) -- (C101) -- cycle;
    \draw[below right] (C000) node {$00(A)$};
    \draw[below right] (C001) node {$11(T)$};
    \draw[below right] (C010) node {$01(C)$};
    \draw[below right] (C011) node {$10(G)$};
    \draw[below right] (C100) node {$10(G)$};
    \draw[below right] (C101) node {$01(C)$};
    \draw[below right] (C110) node {$11(T)$};
    \draw[below right] (C111) node {$00(A)$};
  \end{tikzpicture}
  \caption{Geometrical representation of a linear representation in
    the case of $N = 4$ categories, $k=2$, and $n=3$. Each vertex is
    labeled with a 2-bit string which, as a binary vector, is the
    image of its coordinates under the matrix $\vec{A}$. For
    illustration purpose, each 2-bit string is also arbitrarily
    identified as one of the four nucleobases found in DNA.}
  \label{fig:geometrical_representation}
\end{figure}
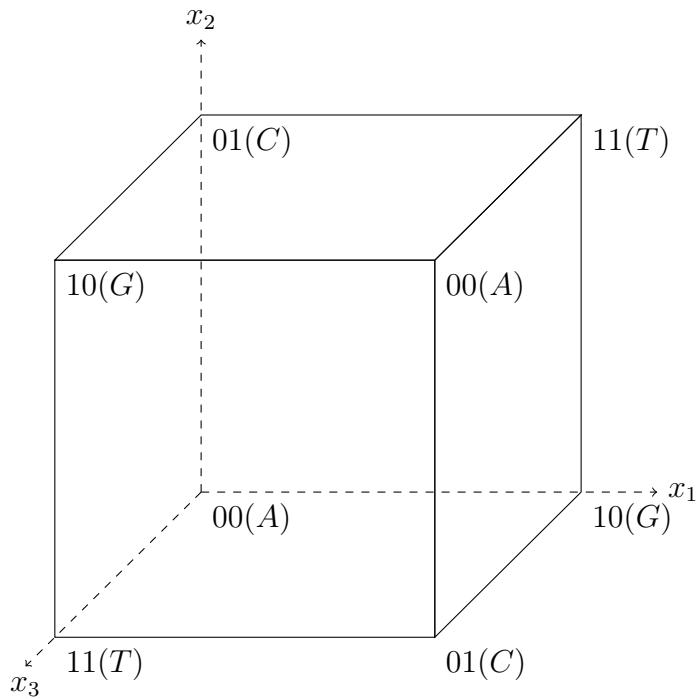

As an example, let us consider again the case of DNA where $N = 4$
categories, $k = 2$, $n = 3$, and
\begin{equation*}
  \vec{A} =
  \begin{pmatrix}
    1 & 0 & 1 \\
    0 & 1 & 1
  \end{pmatrix}
  \,.
\end{equation*}
The order of the columns does not matter.
Fig.~\ref{fig:geometrical_representation} shows a geometric
representation of $\vec{A}$. For example, let $x = (1, 0, 1)^t$, that
is the bottom right vertex on the front face of the cube. Then
\begin{align*}
  \vec{A} \vec{x}
  & =
  \begin{pmatrix}
    1 & 0 & 1 \\
    0 & 1 & 1
  \end{pmatrix}
  \begin{pmatrix}
    1 \\
    0 \\
    1
  \end{pmatrix}
  \\
  & =
  \begin{pmatrix}
    2 \\
    1
  \end{pmatrix}
  =
  \begin{pmatrix}
    0 \\
    1
  \end{pmatrix}
  \pmod{2} \,,
\end{align*}
which gives the bit string $01$ and the nucleobase $C$. Let us compute
the neighbors of $\vec{x}$:
\begin{align*}
  \vec{A}(\vec{x}+\vec{e}_1)
  &= \vec{A}\vec{x} + \vec{A}\vec{e}_1 \\
  &=
    \begin{pmatrix}
      0 \\
      1
    \end{pmatrix}
  +
    \begin{pmatrix}
      1 \\
      0
    \end{pmatrix}
  =
    \begin{pmatrix}
      1 \\
      1
    \end{pmatrix}
  = T \\
  \vec{A}(\vec{x}+\vec{e}_2)
  &= \vec{A}\vec{x} + \vec{A}\vec{e}_2 \\
  &=
    \begin{pmatrix}
      0 \\
      1
    \end{pmatrix}
  +
    \begin{pmatrix}
      0 \\
      1
    \end{pmatrix}
  =
    \begin{pmatrix}
      0 \\
      0
    \end{pmatrix}
  = A \\
  \vec{A}(\vec{x}+\vec{e}_3)
  &= \vec{A}\vec{x} + \vec{A}\vec{e}_3 \\
  &=
    \begin{pmatrix}
      0 \\
      1
    \end{pmatrix}
  +
    \begin{pmatrix}
      1 \\
      1
    \end{pmatrix}
  =
    \begin{pmatrix}
      1 \\
      0
    \end{pmatrix}
  = G
\end{align*}
It is clear that we can reach every categorical value (nucleobase)
other than $C$ in a single mutation.

Following \cite{rothlauf2006representations}, we can say that the
linear representation has high locality; is uniformly redundant (each
categorical value has exactly $2^{n-k}$ representatives); and is non
synonymously redundant (for each categorical value, its
representatives are spread all over the hypercube).

If $N$ is not a power of 2 then we let $k$ be the smallest natural
such that $N < 2^k$ and repeat the construction of $\vec{A}$ with
$n = 2^k - 1$. It is then necessary to map the output of $\vec{A}$ to
$V$ as in
\begin{equation*}
  \vec{x} \mapsto \vec{A}\vec{x} \mapsto i \mapsto j = 1 + (i \mod{N})
  \mapsto v_j \,,
\end{equation*}
where $\vec{A}\vec{x}$ is the binary representation of integer $i$.
The resulting binary representation $\phi$ of $V$ is still locally
surjective but not locally bijective.

From our observations, it seems that the linear representation is
isotropic, a property which generalizes the fact that it is locally
bijective. Intuitively speaking, it means that, from any search point,
a uniform mutation of exactly $r$ bits yields a quasi uniform
distribution over the categorical values. For example, with $N = 16$
categories, $n = 15$ bits, and $r = 3$ bits, from any search point
$\vec{x}$ and its category $\vec{y} = \vec{A}\vec{x}$, there are 35
3-bit mutations yielding the same category $\vec{y}$ and, for each
category $\vec{z}\neq \vec{y}$, there are 28 3-bit mutations yielding
category $\vec{z}$. In total, there are
$35 + 15\times 28 = 455 = \binom{15}{3}$ 3-bit mutations.

For all $\vec{x}\in\binarySet^n$ and all $r\in\range{0}{n}$, let
$S(\vec{x}, r)$ be the Hamming sphere of radius $r$ centered at
$\vec{x}$, that is
$S(\vec{x}, r) = \{\vec{x'}\in\binarySet^n \mid d_H(\vec{x}, \vec{x'})
= r\}$, where $d_H$ is the Hamming distance. We say that a
representation $\phi$ is isotropic if, for all
$\vec{x}\in\binarySet^n$ and $r\in\range{0}{n}$, the map
$\phi : S(\vec{x}, r) \setminus \phi^{-1}(\vec{y}) \to V \setminus
\{\vec{y}\}$, where $\vec{y} = \phi(\vec{x})$, is balanced. In the
case of the linear representation, $\vec{A}$ isotropic is equivalent
to
$\vec{A} : S(0, r) \setminus \ker(\vec{A}) \to \binarySet^k \setminus
\{0\}$ balanced. The question of whether the linear representation is
isotropic remains open.

\section{Experiments}
\label{sec:experiments}

\subsection{Problem}
\label{sec:problem}

We apply linear representation to Sudoku puzzles. A Sudoku puzzle
consists in filling a 9 by 9 board using digits and satisfying a set
of constraints. The board is divided into 3 by 3 blocks. Each row,
column, and block must contain all 1 to 9 digits. We turn a Sudoku
puzzle into an optimization problem by counting the number of
unsatisfied constraints. More precisely, the objective is to minimize
the function $f : V^{81} \to \naturals$ defined by
\begin{align*}
  f(\vec{x}) &= \sum_{i\in V} \sum_{k\in V} \left|\sum_{j\in V}[x_{ij} = k]
               - 1\right| \quad \text{(rows)}\\
             & + \sum_{j\in V} \sum_{k\in V} \left|\sum_{i\in V}[x_{ij} = k]
               - 1\right| \quad \text{(columns)} \\
             & + \sum_{(i, j)\in\{1,4,7\}^2} \sum_{k\in V} \left|\sum_{(k,
               l)\in\{0,1,2\}^2}[x_{i+k,k+l} = k] - 1\right| \quad
               \text{(blocks)} \,,
\end{align*}
where $V = \{1, 2, ..., 9\}$ and $[P] = 1$ if the statement $P$ is
true, 0 otherwise (Iverson bracket). Usually, in a Sudoku puzzle, some
of the $x_{ij}$'s are given. Solving the puzzle is equivalent to
minimizing the function $g$ defined by
$g(\vec{x}_U) = f(\vec{x}_U, \vec{x}_K)$, where $U \subset V \times V$
is the set of locations of unknowns and $K = (V \times V) \setminus U$
is the set of locations of known digits. The puzzle is solvable if and
only if the minimum of $g$ is zero.

From the point of view of linear representation, Sudoku is a
worst-case scenario since $N = 9$ is one past a power of 2. Direct
representation requires $n = 4$ bits whereas linear representation
requires $n=15$ bits. For example, instance sudoku-intermediate-54 has
54 unknowns and the size of its (binary) search space is 216 bits with
direct representation and 810 bits with linear representation.

\subsection{Fixed-budget experiments}
\label{sec:fixed-budg-exper}

We have generated Sudoku instances of varying difficulty using an
online generator (which is also open source)
\cite{ostermiller:_qqwin}. It should be noted that the difficulty
grade (simple, easy, intermediate, expert) is relevant to algorithms
relying on constraint programming or backtracking rather than
metaheuristics. We have generated two instances per difficulty grade.
We have also included two 17-hint uniquely completable instances
\cite{royle:_minim_sudok}.

The study includes the following metaheuristics: random local search
with restart (RLS), hill climbing with restart (HC), simulated
annealing (SA) \cite{kirkpatrick-gelatt-vecchi-1983}, genetic
algorithm (GA) \cite{holland-75}, $(1+1)$ evolutionary algorithm (EA),
$(10+1)$ evolutionary algorithm, population-based incremental learning
(PBIL) \cite{baluja-caruana-95}, mutual information maximization for
input clustering (MIMIC) \cite{isbell}, univariate marginal
distribution algorithm (UMDA) \cite{muhlenbein-97}, linkage tree
genetic algorithm (LTGA) \cite{thierens-2010}, parameter-less
population pyramid (P3) \cite{goldman-punch-2015}. All metaheuristics
have been applied to the same set of instances with direct and linear
representations. They have been given the same arbitrary budget
(\np{300000} function evaluations) and run 20 times. All experiments
have been produced with the HNCO framework \cite{hnco}.

\begin{figure}
  \centering
  \includegraphics[width=0.8\linewidth]{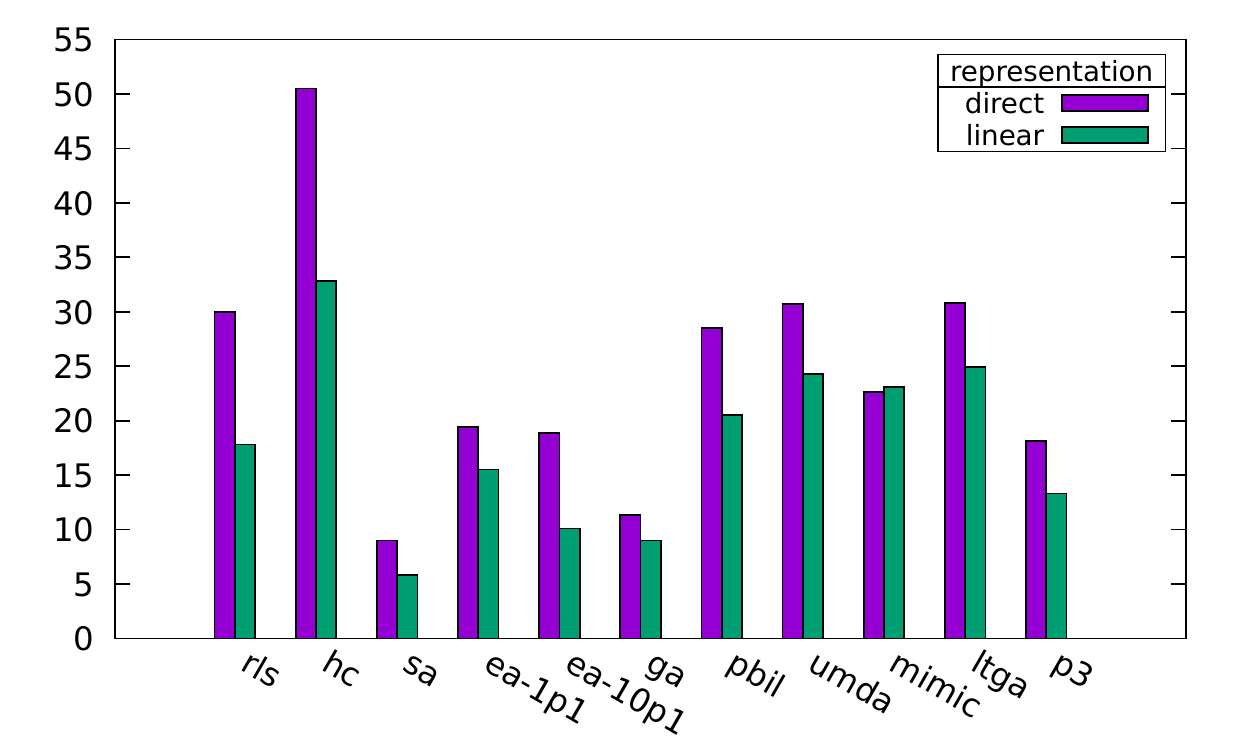}
  \caption{Mean value of solutions found by metaheuristics with direct
    and linear representations (lower is better, instance
    sudoku-intermediate-54, \np{300000} function evaluations, 20
    runs).}
  \label{fig:fixed_budget_histogram}
\end{figure}

Fig.~\ref{fig:fixed_budget_histogram} shows the mean values of
solutions found by metaheuristics on a particular instance
(sudoku-intermediate-54). Despite increased dimension of the search
space, all metaheuristics but MIMIC consistently improve the quality
of their solutions with linear representation.

Tab.~\ref{tab:fixed_budget_value_statistics} shows the summary
statistics of solutions on instance sudoku-easy-56. Metaheuristics are
ranked according to their median, first and third quartiles, minimum
and maximum, in that order. On this particular instance, only MIMIC
does not take advantage of linear representation.

Tab.~\ref{tab:fixed_budget_rank_distributions} shows the summary
statistics of ranks on the set of all ten instances. Metaheuristics
are globally ranked. On this particular set of instances and within
the given budget, for all metaheuristics, linear representation
dominates direct representation.

\begin{table}
  \centering
  \begin{tabular}{@{} ll >{{\nprounddigits{0}}}n{2}{0}>{{\nprounddigits{2}}}n{2}{2}>{{\nprounddigits{1}}}n{2}{1}>{{\nprounddigits{2}}}n{2}{2}>{{\nprounddigits{0}}}n{2}{0} @{}}
\toprule
{Algo.} & {rep.} & \multicolumn{5}{l}{{Value}} \\
\midrule
&& {min} & {$Q_1$} & {med.} & {$Q_3$} & {max} \\
\midrule
sa & linear & 4.000000 & 4.000000 & 4.000000 & 6.000000 & 8.000000 \\
sa & direct & 4.000000 & 8.000000 & 8.000000 & 12.500000 & 14.000000 \\
ga & linear & 4.000000 & 8.000000 & 10.000000 & 12.000000 & 16.000000 \\
ea-10p1 & linear & 6.000000 & 9.500000 & 10.000000 & 12.000000 & 18.000000 \\
ga & direct & 4.000000 & 9.500000 & 12.000000 & 14.000000 & 18.000000 \\
p3 & linear & 10.000000 & 13.500000 & 14.000000 & 16.000000 & 18.000000 \\
ea-1p1 & linear & 8.000000 & 13.500000 & 17.000000 & 18.000000 & 24.000000 \\
rls & linear & 12.000000 & 18.000000 & 19.000000 & 20.000000 & 24.000000 \\
p3 & direct & 14.000000 & 18.000000 & 21.000000 & 22.000000 & 28.000000 \\
ea-10p1 & direct & 10.000000 & 18.000000 & 22.000000 & 24.500000 & 30.000000 \\
umda & linear & 16.000000 & 20.000000 & 22.000000 & 26.000000 & 28.000000 \\
pbil & linear & 14.000000 & 20.000000 & 23.000000 & 26.000000 & 32.000000 \\
ea-1p1 & direct & 12.000000 & 21.500000 & 24.000000 & 26.500000 & 32.000000 \\
mimic & direct & 16.000000 & 22.000000 & 24.000000 & 26.000000 & 34.000000 \\
ltga & linear & 22.000000 & 24.000000 & 25.000000 & 26.500000 & 30.000000 \\
mimic & linear & 20.000000 & 25.500000 & 28.000000 & 28.500000 & 38.000000 \\
pbil & direct & 26.000000 & 31.000000 & 32.000000 & 36.000000 & 38.000000 \\
umda & direct & 26.000000 & 28.000000 & 33.000000 & 36.000000 & 42.000000 \\
hc & linear & 28.000000 & 31.500000 & 36.000000 & 36.500000 & 42.000000 \\
rls & direct & 28.000000 & 34.000000 & 36.000000 & 36.000000 & 38.000000 \\
ltga & direct & 30.000000 & 35.500000 & 36.000000 & 38.000000 & 40.000000 \\
hc & direct & 42.000000 & 49.500000 & 52.000000 & 52.000000 & 56.000000 \\
\bottomrule
\end{tabular}

  \caption{Value statistics in fixed-budget experiments (lower is
    better, instance sudoku-easy-56, \np{300000} function evaluations,
    20 runs).}
  \label{tab:fixed_budget_value_statistics}
\end{table}

\begin{table}
  \centering
  \begin{tabular}{@{} ll >{{\nprounddigits{0}}}n{2}{0}>{{\nprounddigits{2}}}n{2}{2}>{{\nprounddigits{1}}}n{2}{1}>{{\nprounddigits{2}}}n{2}{2}>{{\nprounddigits{0}}}n{2}{0} @{}}
\toprule
{Algo.} & {rep.} & \multicolumn{5}{l}{{Rank}} \\
\midrule
&& {min} & {$Q_1$} & {med.} & {$Q_3$} & {max} \\
\midrule
sa & linear & 1.000000 & 1.000000 & 1.000000 & 1.000000 & 1.000000 \\
ga & linear & 2.000000 & 2.000000 & 2.000000 & 3.000000 & 3.000000 \\
sa & direct & 2.000000 & 2.000000 & 2.500000 & 3.000000 & 3.000000 \\
ea-10p1 & linear & 4.000000 & 4.000000 & 4.000000 & 4.750000 & 5.000000 \\
ga & direct & 4.000000 & 4.250000 & 5.000000 & 5.000000 & 6.000000 \\
ea-1p1 & linear & 5.000000 & 6.000000 & 6.000000 & 7.000000 & 7.000000 \\
p3 & linear & 6.000000 & 6.000000 & 7.000000 & 7.000000 & 7.000000 \\
ea-10p1 & direct & 8.000000 & 8.000000 & 8.500000 & 9.750000 & 10.000000 \\
rls & linear & 8.000000 & 8.000000 & 9.000000 & 9.000000 & 10.000000 \\
p3 & direct & 8.000000 & 9.000000 & 9.500000 & 10.750000 & 15.000000 \\
ea-1p1 & direct & 9.000000 & 10.250000 & 11.000000 & 11.750000 & 13.000000 \\
pbil & linear & 11.000000 & 11.250000 & 12.000000 & 12.000000 & 15.000000 \\
mimic & linear & 12.000000 & 13.000000 & 13.500000 & 14.750000 & 16.000000 \\
umda & linear & 11.000000 & 13.000000 & 14.000000 & 14.000000 & 15.000000 \\
mimic & direct & 11.000000 & 13.250000 & 14.000000 & 15.000000 & 15.000000 \\
ltga & linear & 15.000000 & 16.000000 & 16.000000 & 16.000000 & 16.000000 \\
pbil & direct & 17.000000 & 17.000000 & 17.000000 & 17.000000 & 18.000000 \\
umda & direct & 17.000000 & 18.000000 & 18.500000 & 19.000000 & 20.000000 \\
rls & direct & 18.000000 & 19.000000 & 19.000000 & 20.000000 & 21.000000 \\
hc & linear & 18.000000 & 18.250000 & 20.000000 & 20.000000 & 21.000000 \\
ltga & direct & 19.000000 & 20.000000 & 21.000000 & 21.000000 & 21.000000 \\
hc & direct & 22.000000 & 22.000000 & 22.000000 & 22.000000 & 22.000000 \\
\bottomrule
\end{tabular}

  \caption{Rank statistics in fixed-budget experiments (value based,
    all ten instances).}
  \label{tab:fixed_budget_rank_distributions}
\end{table}

\subsection{Empirical cumulative distribution functions}
\label{sec:empir-cumul-distr}

To account for the dynamical behavior of metaheuristics with respect
to representation, we have studied their empirical cumulative
distribution functions (ECDF) \cite{DBLP:journals/corr/HansenABTT16}.
For each instance, every metaheuristic has been run 20 times with a
budget of $10^7$ evaluations per run. For each run, every improvement
has been recorded. Then, for each instance, the range of function
values has been evenly divided into 50 targets. Finally, for each
metaheuristic, the mean proportion of targets reached at each number
of evaluations has been computed.

\begin{figure}
  \centering
  \includegraphics[width=0.8\linewidth]{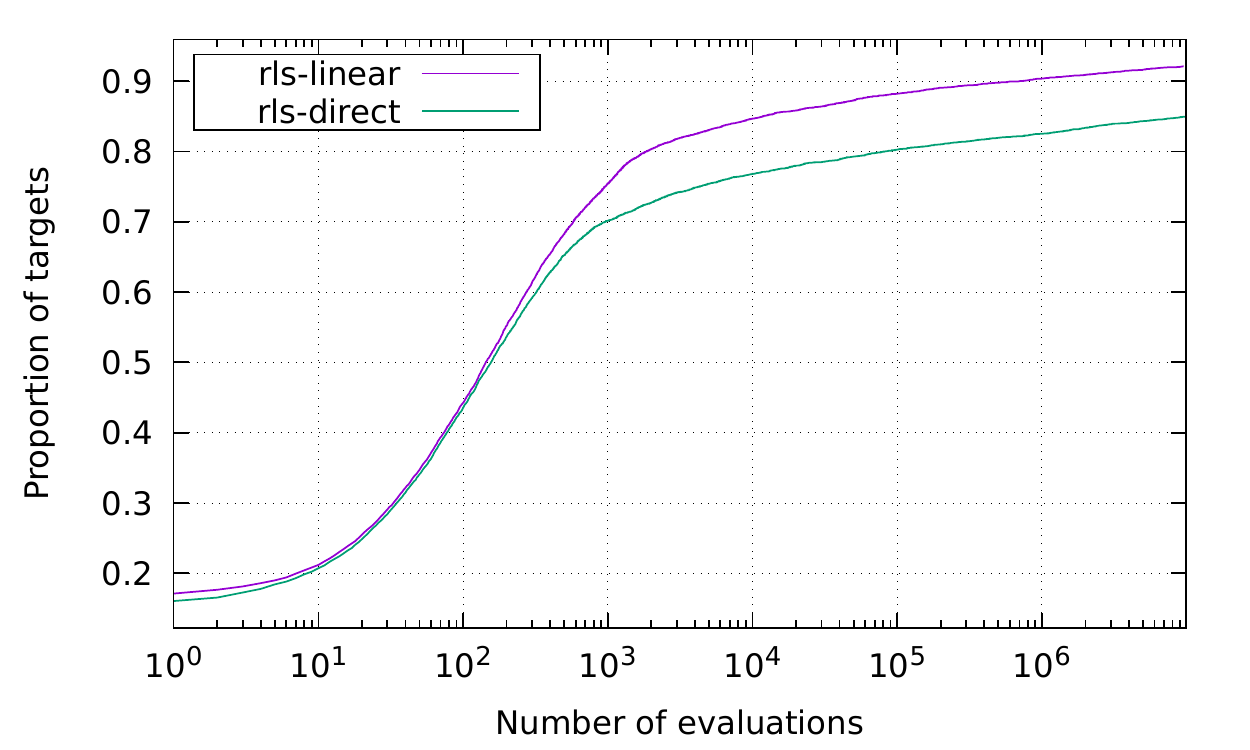}
  \caption{ECDF's of RLS with direct and linear representations (20
    runs).}
  \label{fig:ecdf_rls}
\end{figure}

\begin{figure}
  \centering
  \includegraphics[width=0.8\linewidth]{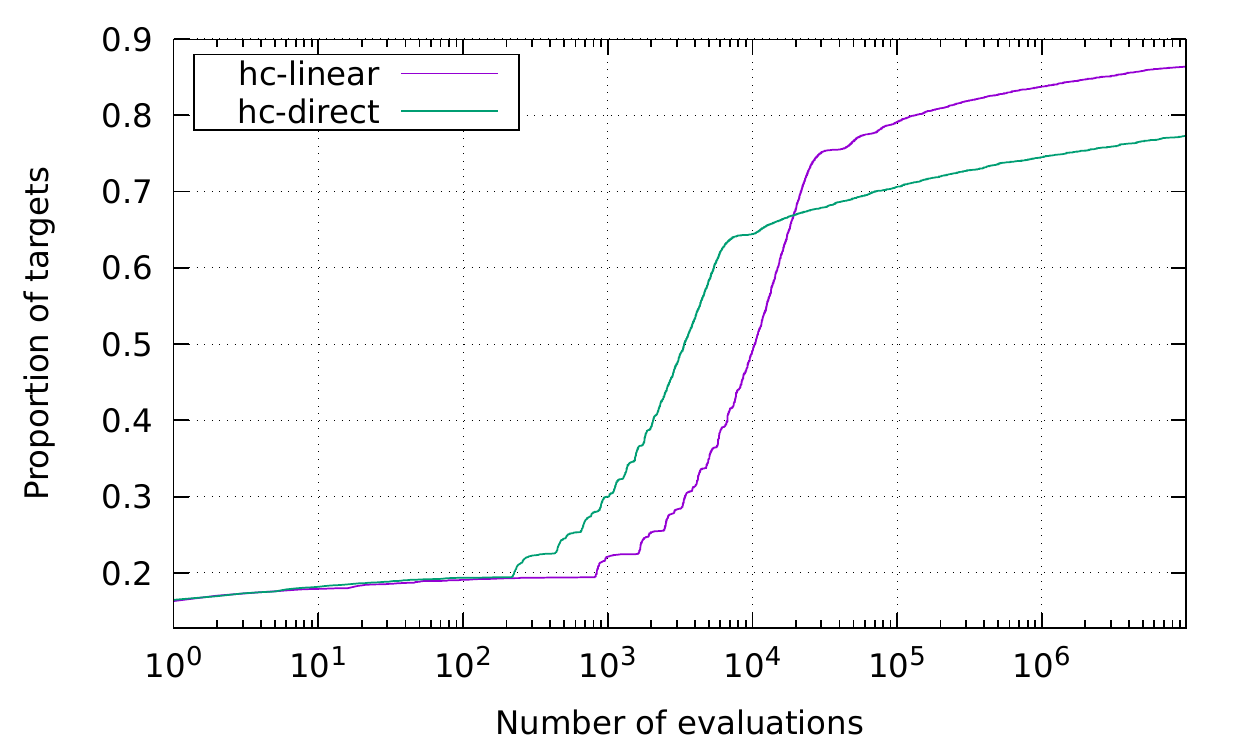}
  \caption{ECDF's of HC with direct and linear representations (20
    runs).}
  \label{fig:ecdf_hc}
\end{figure}

\begin{figure}
  \centering
  \includegraphics[width=0.8\linewidth]{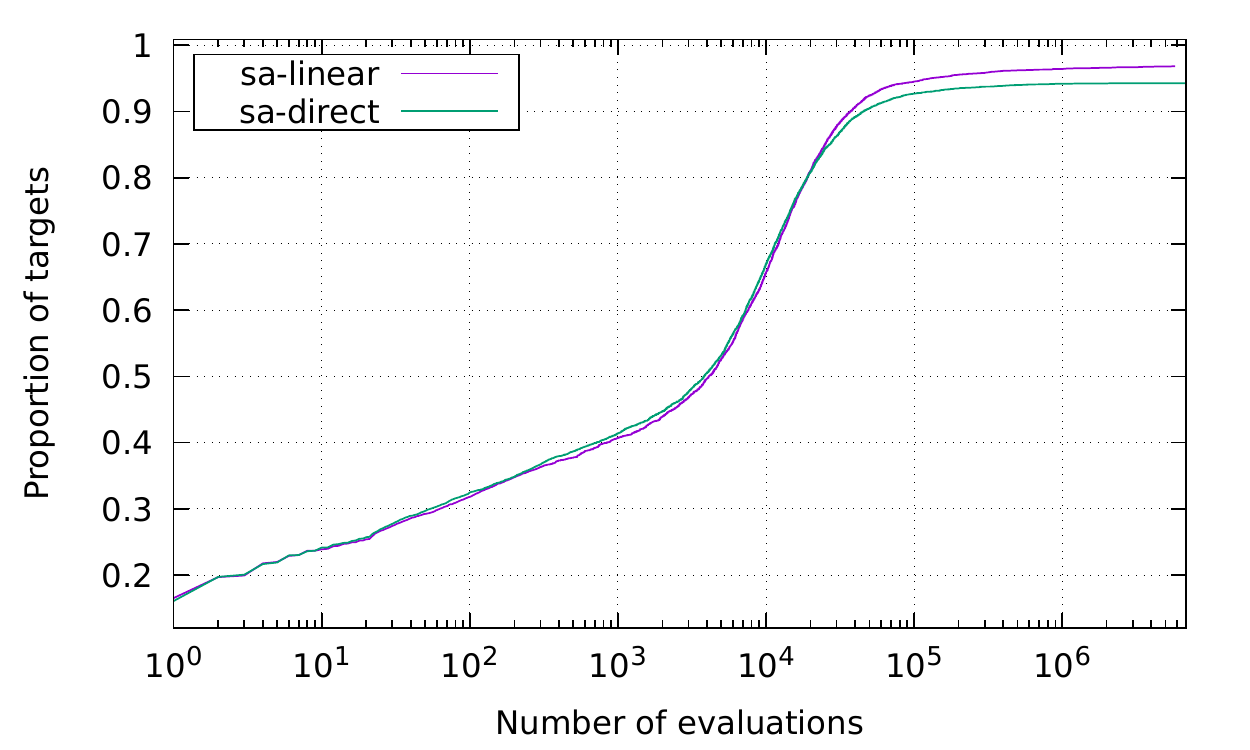}
  \caption{ECDF's of SA with direct and linear representations (20
    runs).}
  \label{fig:ecdf_sa}
\end{figure}

\begin{figure}
  \centering
  \includegraphics[width=0.8\linewidth]{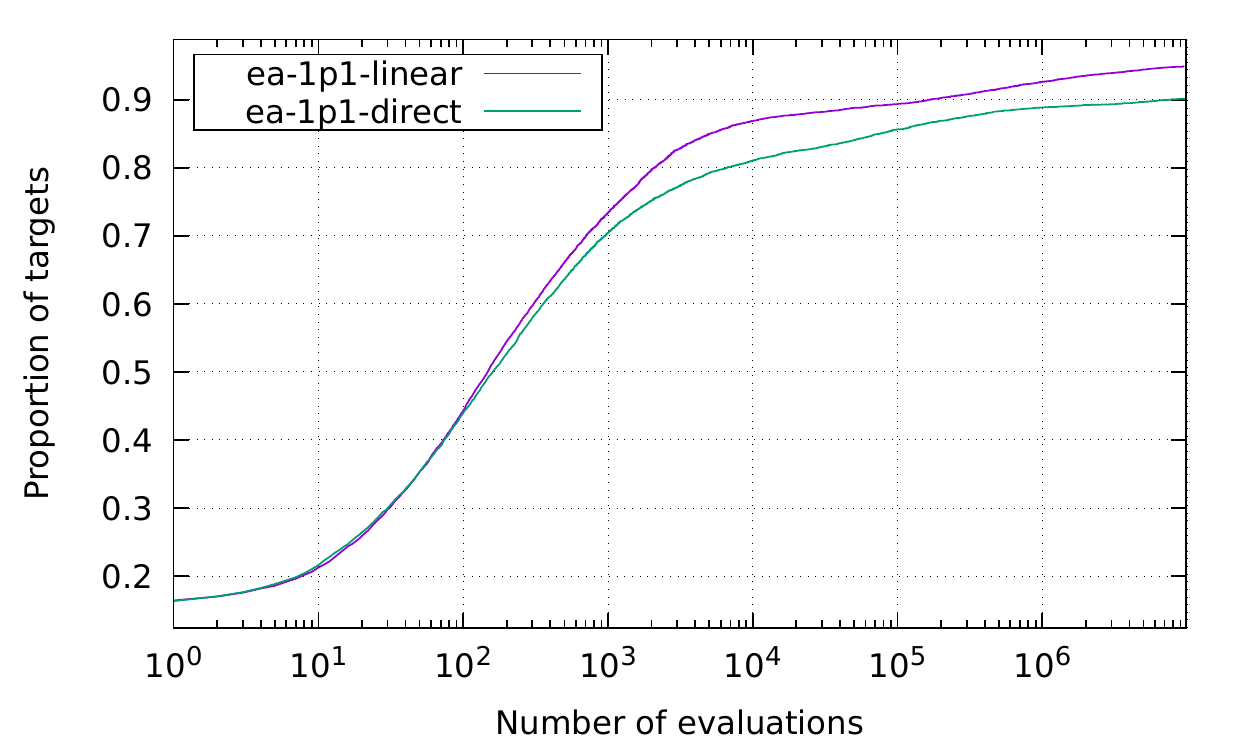}
  \caption{ECDF's of $(1+1)$ EA with direct and linear representations
    (20 runs).}
  \label{fig:ecdf_ea_1p1}
\end{figure}

\begin{figure}
  \centering
  \includegraphics[width=0.8\linewidth]{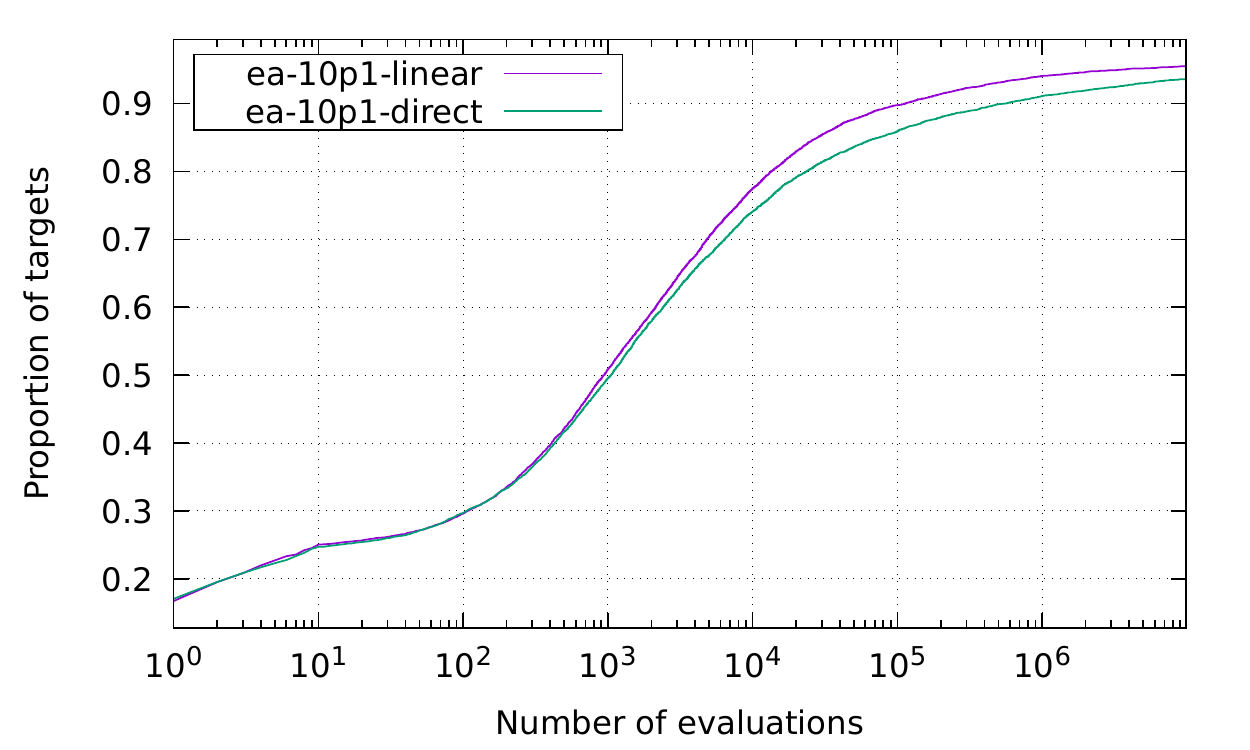}
  \caption{ECDF's of $(10+1)$ EA with direct and linear
    representations (20 runs).}
  \label{fig:ecdf_ea_10p1}
\end{figure}

\begin{figure}
  \centering
  \includegraphics[width=0.8\linewidth]{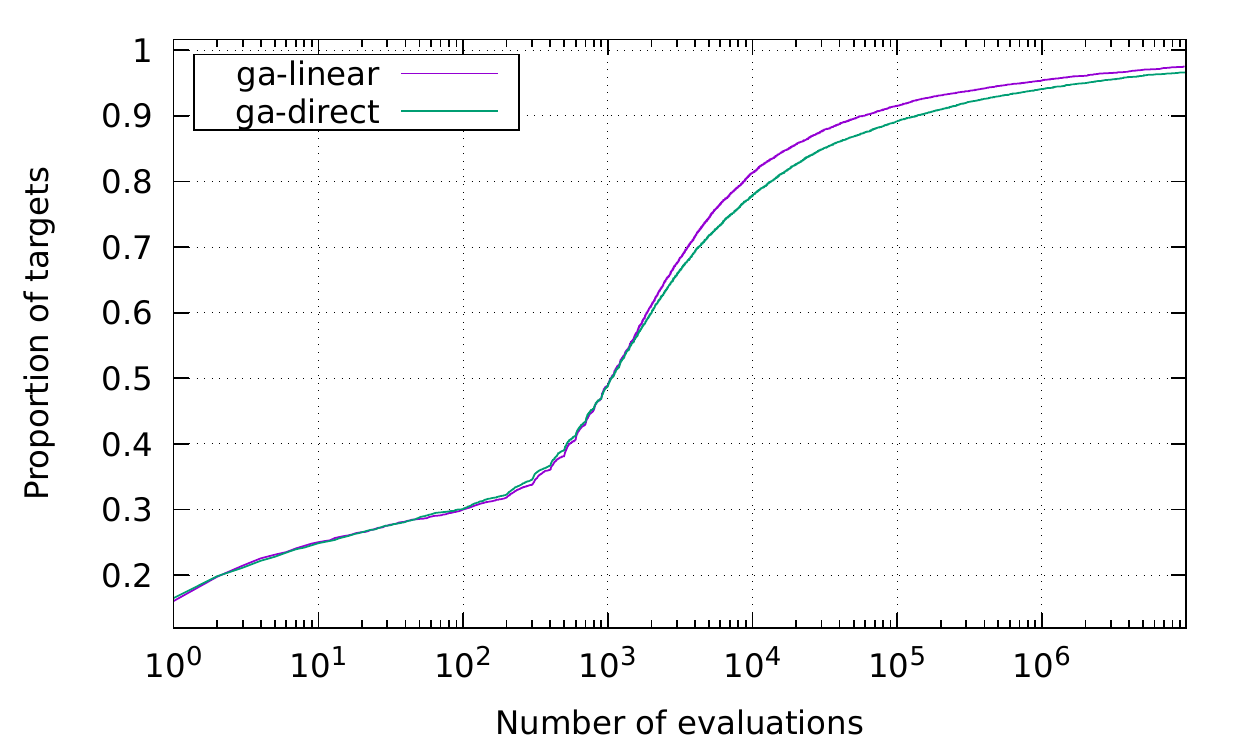}
  \caption{ECDF's of GA with direct and linear representations (20
    runs).}
  \label{fig:ecdf_ga}
\end{figure}

\begin{figure}
  \centering
  \includegraphics[width=0.8\linewidth]{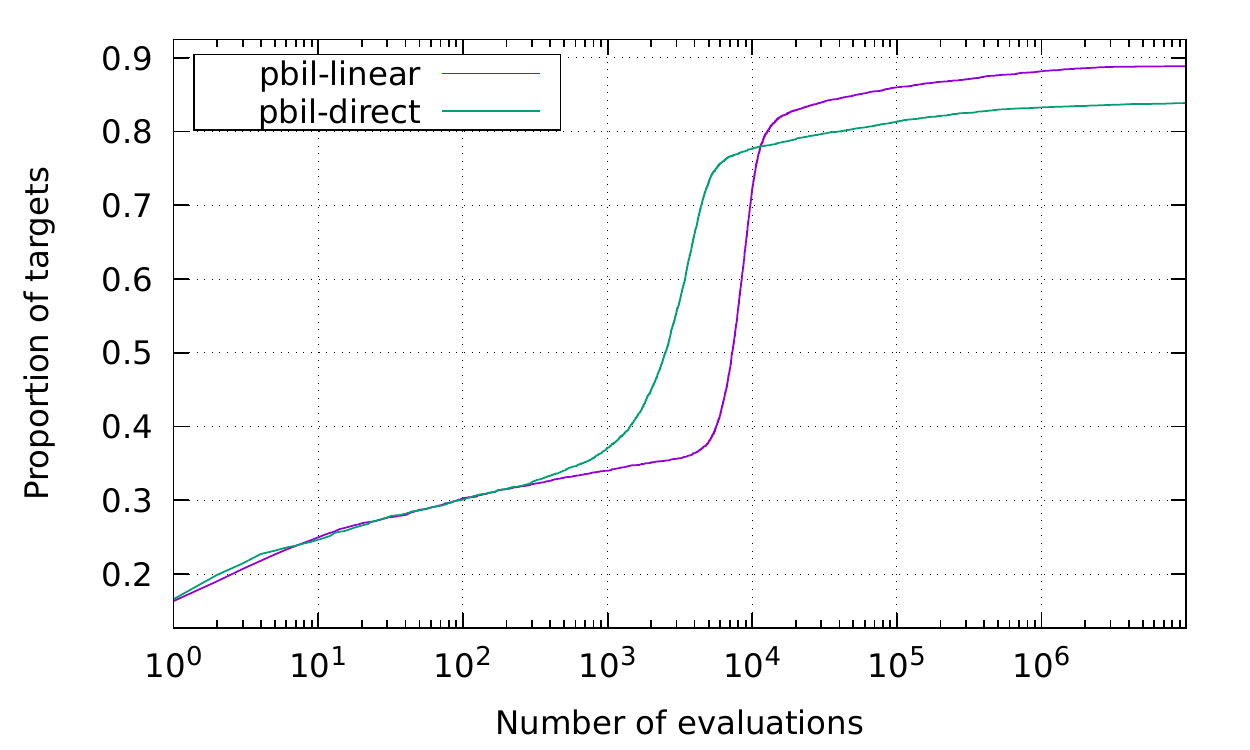}
  \caption{ECDF's of PBIL with direct and linear representations (20
    runs).}
  \label{fig:ecdf_pbil}
\end{figure}

\begin{figure}
  \centering
  \includegraphics[width=0.8\linewidth]{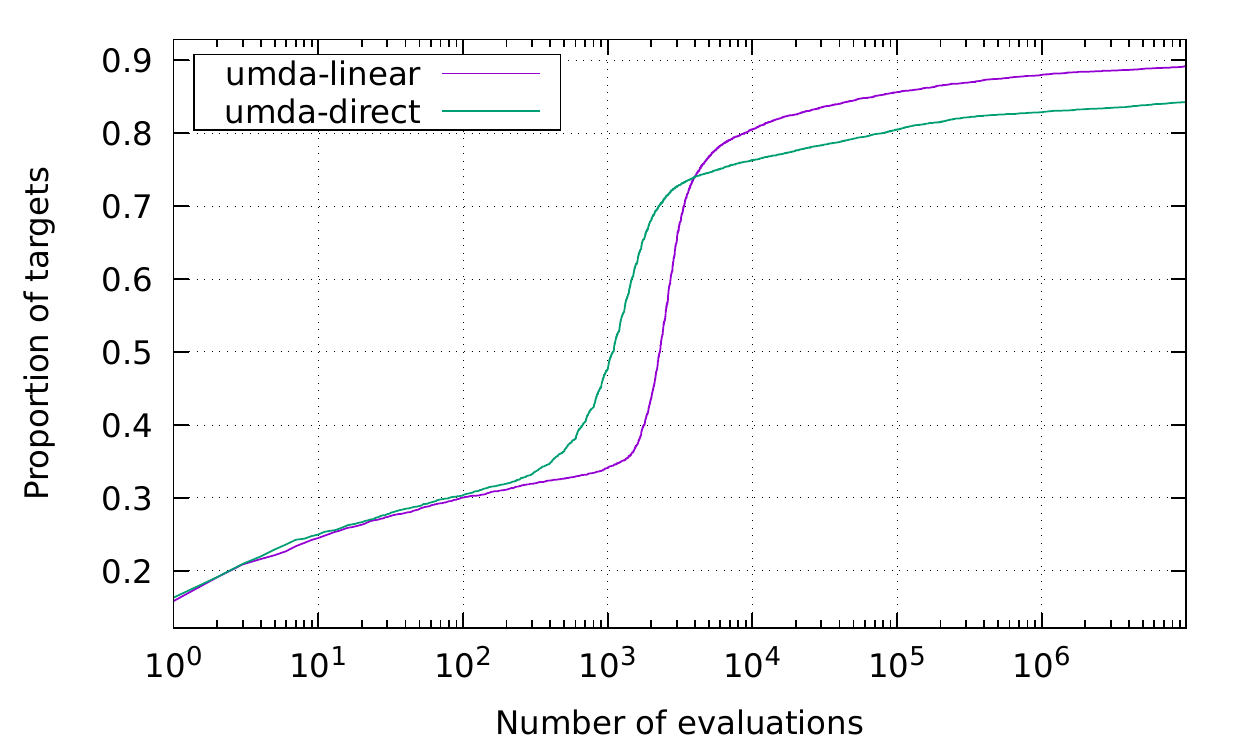}
  \caption{ECDF's of UMDA with direct and linear representations (20
    runs).}
  \label{fig:ecdf_umda}
\end{figure}

\begin{figure}
  \centering
  \includegraphics[width=0.8\linewidth]{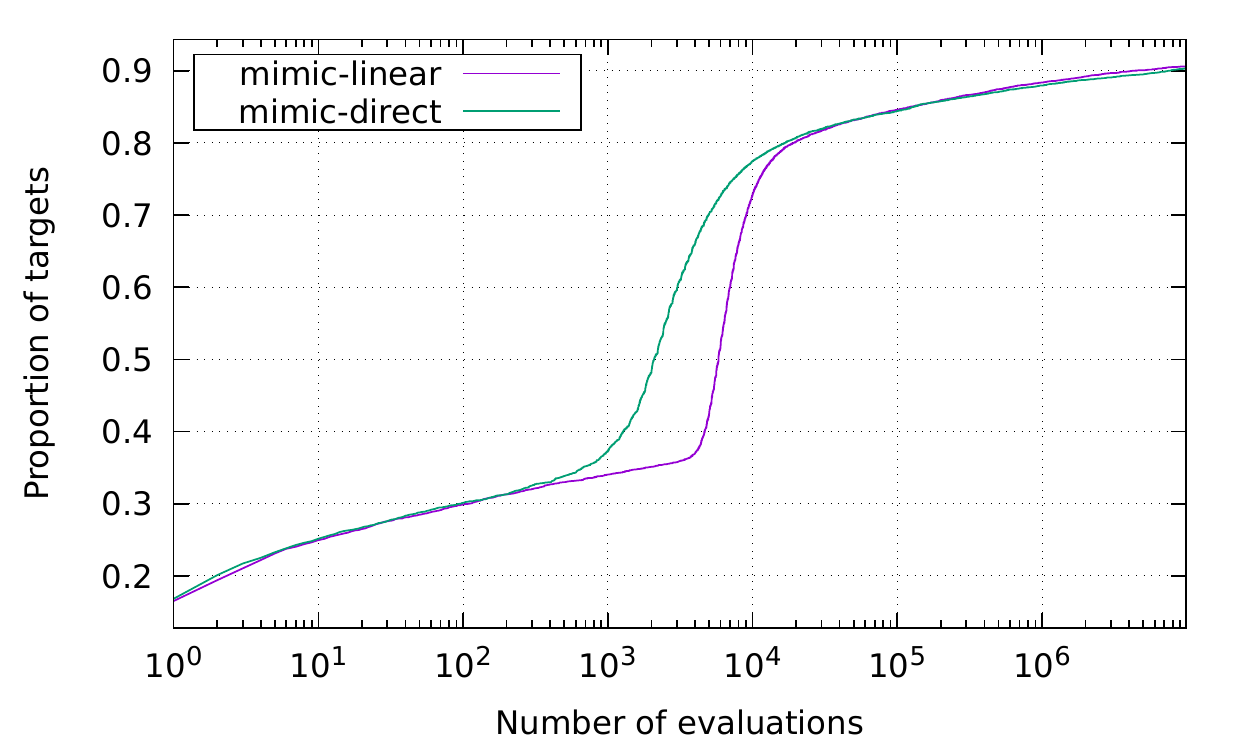}
  \caption{ECDF's of MIMIC with direct and linear representations (20
    runs).}
  \label{fig:ecdf_mimic}
\end{figure}

\begin{figure}
  \centering
  \includegraphics[width=0.8\linewidth]{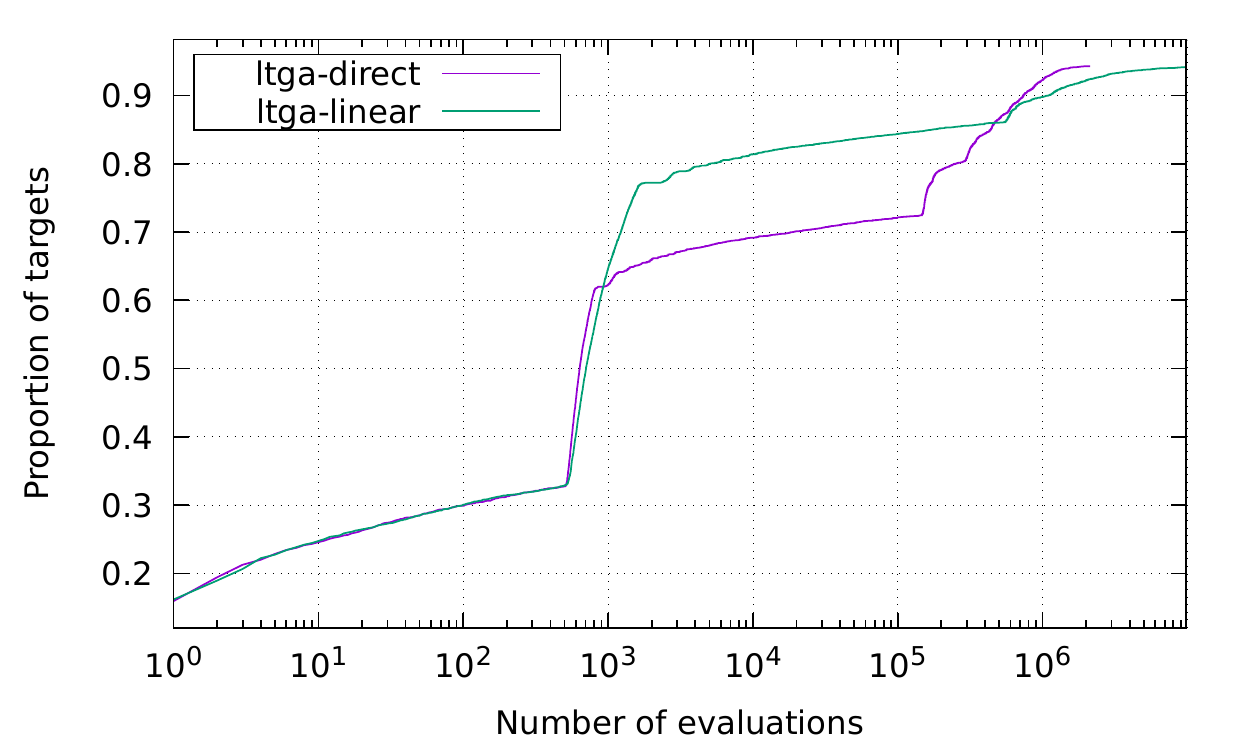}
  \caption{ECDF's of LTGA with direct and linear representations (20
    runs).}
  \label{fig:ecdf_ltga}
\end{figure}

\begin{figure}
  \centering
  \includegraphics[width=0.8\linewidth]{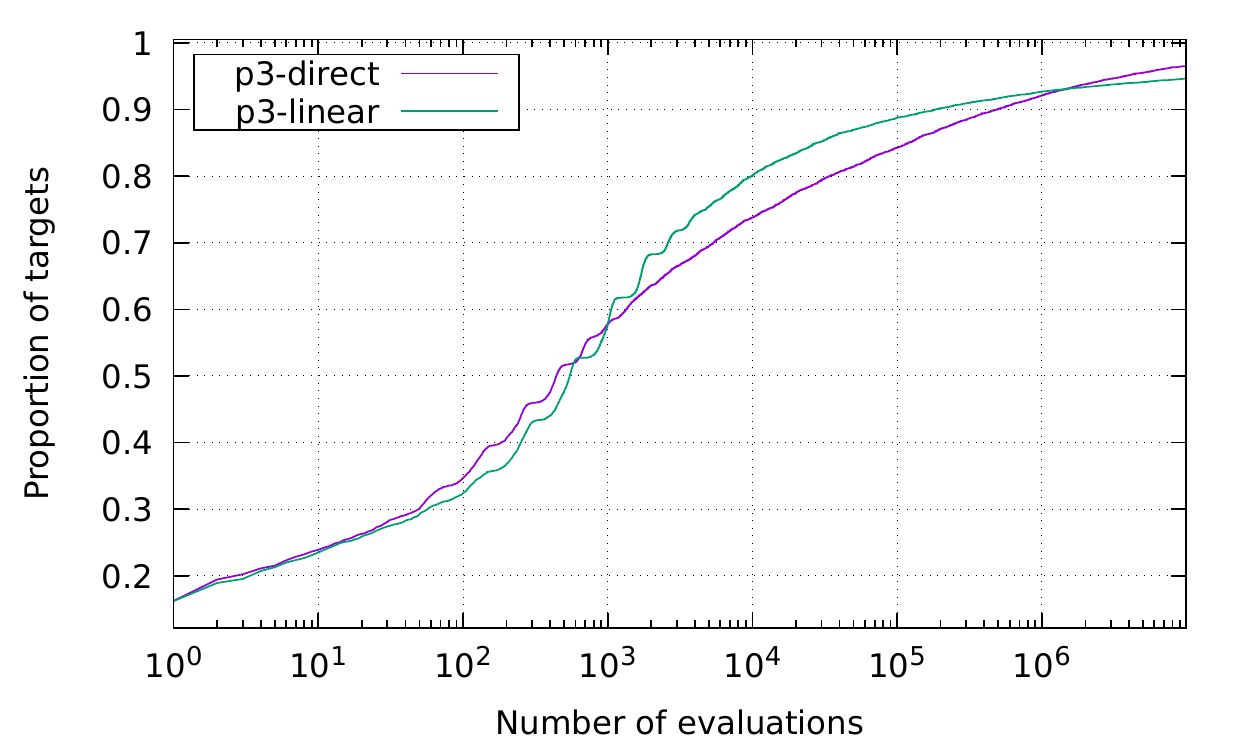}
  \caption{ECDF's of P3 with direct and linear representations (20
    runs).}
  \label{fig:ecdf_p3}
\end{figure}

The metaheuristics and instances considered in this section are the
same as in the previous one. Result are shown in
Fig.~\ref{fig:ecdf_rls}-\ref{fig:ecdf_p3}. It should be noted that
representations, linear or direct, are ranked in the keys according to
their final scores. If a curve stops before $10^7$ evaluations, it
means that the metaheuristic-representation pair did not make any
further progress. Almost all curves have the shape of a sigmoid. We
can identity the following patterns:
\begin{itemize}

\item The curves gradually diverge and linear representation
  dominates. This is the case for RLS (Fig.~\ref{fig:ecdf_rls}), SA
  (Fig.~\ref{fig:ecdf_sa}), $(1+1)$ EA (Fig.~\ref{fig:ecdf_ea_1p1}),
  $(10+1)$ EA (Fig.~\ref{fig:ecdf_ea_10p1}), and GA
  (Fig.~\ref{fig:ecdf_ga}).

\item The curve for linear representation eventually crosses from
  below the one for direct representation after a delayed and sharp
  transition. This is the case for HC (Fig.~\ref{fig:ecdf_hc}), PBIL
  (Fig.~\ref{fig:ecdf_pbil}), UMDA (Fig.~\ref{fig:ecdf_umda}), and, by
  a small margin, MIMIC (Fig.~\ref{fig:ecdf_mimic}).

\end{itemize}
In the case of P3 (Fig.~\ref{fig:ecdf_p3}), the curves have a smooth
staircase shape before a few thousands evaluations but can still be
considered as sigmoids on a larger scale. Linear representation
dominates in the range $[10^3, 1.5 \cdot 10^6]$ and direct
representation overtakes it afterward. In the case of LTGA
(Fig.~\ref{fig:ecdf_ltga}), the curves do not have the shape of a
sigmoid. Linear representation dominates in the range
$[924, 4.7\cdot 10^5]$ and direct representation overtakes it
afterward. However, LTGA with linear representation makes progress
until the end, on the contrary to LTGA with direct representation, and
almost catches up with it.

\subsection{Fixed-target experiments}
\label{sec:fixed-targ-exper}

As we have seen in fixed-budget experiments, metaheuristics almost
never find an optimal solution, that is a solution which satisfies
every constraint. As a consequence, we have generated easy instances,
starting from complete boards and erasing a small number
$r\in\range{1}{10}$ of digits. For each dimension, 4 instances have
been generated. For each instance, every metaheuristic has been run
100 times. A run is successful if an optimal solution has been found
before $10^6$ evaluations. In this case, the runtime is the number of
evaluations needed to find it. For each instance, metaheuristics are
ranked according to median runtime. Finally, they are ranked on a
given set of instances according to success rate then rank statistics.
It should be noted that all runs until $r = 6$ erased digits are
successful. Hence, the full set of instances has been further divided
into low dimension instances, for which $r\in\range{1}{5}$, and medium
dimension instances, for which $r\in\range{6}{10}$.
Tab.~\ref{tab:fixed_target_summary_statistics_3} gives the results for
a particular small dimension instance (small-dimension-3-a) with
$r = 3$ missing digits.
Tab.~\ref{tab:fixed_target_summary_statistics_8} gives the results for
a particular medium dimension instance (small-dimension-8-a) with
$r = 8$ missing digits. Representation can have a significant impact
on maximum runtime, as exemplified in this table by UMDA, MIMIC, HC,
and $(1+1)$ EA.

\begin{table}
  \centering
  \begin{tabular}{@{} l >{{\nprounddigits{0}}}n{4}{0}>{{\nprounddigits{2}}}n{4}{2}>{{\nprounddigits{1}}}n{4}{1}>{{\nprounddigits{2}}}n{4}{2}>{{\nprounddigits{0}}}n{4}{0} >{{\nprounddigits{1} \npunit{\%}}}n{3}{1} @{}}
\toprule
{Algorithm} & \multicolumn{5}{l}{{Number of evaluations}} & {Success} \\
\midrule
& {min} & {$Q_1$} & {med.} & {$Q_3$} & {max} & \\
\midrule
rls-linear & 6.000000 & 18.750000 & 30.000000 & 50.250000 & 269.000000 & 100\\
rls-direct & 1.000000 & 20.000000 & 38.000000 & 64.500000 & 600.000000 & 100\\
ea-1p1-linear & 4.000000 & 29.000000 & 45.000000 & 71.000000 & 192.000000 & 100\\
ea-1p1-direct & 3.000000 & 28.500000 & 47.500000 & 72.750000 & 336.000000 & 100\\
p3-linear & 5.000000 & 56.750000 & 87.000000 & 134.250000 & 316.000000 & 100\\
sa-linear & 3.000000 & 46.250000 & 93.500000 & 216.500000 & 1171.000000 & 100\\
hc-linear & 1.000000 & 79.500000 & 97.500000 & 113.000000 & 128.000000 & 100\\
p3-direct & 11.000000 & 67.500000 & 107.000000 & 157.500000 & 357.000000 & 100\\
hc-direct & 1.000000 & 59.250000 & 115.000000 & 248.000000 & 1155.000000 & 100\\
umda-direct & 3.000000 & 118.250000 & 156.500000 & 207.500000 & 348.000000 & 100\\
ga-direct & 2.000000 & 109.000000 & 166.000000 & 217.750000 & 415.000000 & 100\\
ga-linear & 4.000000 & 111.750000 & 181.500000 & 238.500000 & 419.000000 & 100\\
pbil-direct & 2.000000 & 127.000000 & 186.500000 & 265.000000 & 461.000000 & 100\\
mimic-direct & 1.000000 & 136.750000 & 209.500000 & 244.250000 & 418.000000 & 100\\
ea-10p1-direct & 2.000000 & 107.000000 & 228.000000 & 365.500000 & 1151.000000 & 100\\
ea-10p1-linear & 4.000000 & 107.000000 & 269.000000 & 430.000000 & 1791.000000 & 100\\
pbil-linear & 13.000000 & 209.000000 & 307.000000 & 658.750000 & 1858.000000 & 100\\
umda-linear & 1.000000 & 104.250000 & 348.000000 & 715.000000 & 1566.000000 & 100\\
ltga-linear & 2.000000 & 135.000000 & 352.500000 & 524.500000 & 610.000000 & 100\\
mimic-linear & 2.000000 & 130.500000 & 387.500000 & 686.250000 & 1715.000000 & 100\\
ltga-direct & 5.000000 & 140.750000 & 443.500000 & 555.250000 & 840.000000 & 100\\
sa-direct & 13.000000 & 39.500000 & 102.000000 & 243.750000 & 845.000000 & 99\\
\bottomrule
\end{tabular}

  \caption{Runtime statistics in fixed-target experiments (instance
    small-dimension-3-a, 100 runs).}
  \label{tab:fixed_target_summary_statistics_3}
\end{table}

\begin{table}
  \centering
  \begin{tabular}{@{} l >{{\nprounddigits{0}}}n{6}{0}>{{\nprounddigits{2}}}n{6}{2}>{{\nprounddigits{1}}}n{6}{1}>{{\nprounddigits{2}}}n{6}{2}>{{\nprounddigits{0}}}n{6}{0} >{{\nprounddigits{1} \npunit{\%}}}n{3}{1} @{}}
\toprule
{Algorithm} & \multicolumn{5}{l}{{Number of evaluations}} & {Success} \\
\midrule
& {min} & {$Q_1$} & {med.} & {$Q_3$} & {max} & \\
\midrule
rls-linear & 46.000000 & 154.000000 & 259.000000 & 815.500000 & 4217.000000 & 100\\
ea-1p1-linear & 50.000000 & 169.000000 & 266.000000 & 536.750000 & 19447.000000 & 100\\
rls-direct & 38.000000 & 162.750000 & 403.000000 & 834.250000 & 3920.000000 & 100\\
p3-linear & 141.000000 & 473.000000 & 709.500000 & 992.250000 & 1979.000000 & 100\\
p3-direct & 275.000000 & 513.500000 & 782.500000 & 1092.250000 & 2095.000000 & 100\\
mimic-direct & 432.000000 & 801.250000 & 861.500000 & 1033.250000 & 16391.000000 & 100\\
hc-linear & 532.000000 & 757.000000 & 876.000000 & 1657.500000 & 4481.000000 & 100\\
ltga-linear & 567.000000 & 641.750000 & 920.000000 & 1112.000000 & 2218.000000 & 100\\
ga-direct & 302.000000 & 813.250000 & 994.000000 & 1277.250000 & 2010.000000 & 100\\
ga-linear & 510.000000 & 962.250000 & 1228.000000 & 1451.750000 & 2956.000000 & 100\\
ea-10p1-linear & 432.000000 & 955.750000 & 1569.000000 & 2383.000000 & 5572.000000 & 100\\
ea-10p1-direct & 422.000000 & 1078.000000 & 1696.500000 & 2554.000000 & 8081.000000 & 100\\
umda-linear & 1351.000000 & 1770.500000 & 1948.500000 & 2171.000000 & 20398.000000 & 100\\
mimic-linear & 2519.000000 & 3663.500000 & 4139.500000 & 4701.250000 & 25444.000000 & 100\\
ltga-direct & 608.000000 & 2047.750000 & 4151.500000 & 6835.500000 & 21157.000000 & 100\\
pbil-linear & 3567.000000 & 4823.250000 & 5160.500000 & 5647.250000 & 9616.000000 & 100\\
hc-direct & 920.000000 & 9232.750000 & 22709.500000 & 39932.500000 & 166010.000000 & 100\\
umda-direct & 213.000000 & 519.750000 & 574.500000 & 655.250000 & 922799.000000 & 98\\
sa-direct & 167.000000 & 4675.750000 & 6833.500000 & 8244.500000 & 12039.000000 & 98\\
sa-linear & 101.000000 & 4082.500000 & 6878.000000 & 8872.000000 & 13342.000000 & 98\\
pbil-direct & 746.000000 & 905.500000 & 1029.000000 & 1174.750000 & 3315.000000 & 94\\
ea-1p1-direct & 55.000000 & 188.250000 & 317.000000 & 715.750000 & 949772.000000 & 90\\
\bottomrule
\end{tabular}

  \caption{Runtime statistics in fixed-target experiments (instance
    small-dimension-8-a, 100 runs).}
  \label{tab:fixed_target_summary_statistics_8}
\end{table}

Tab.~\ref{tab:fixed_target_rank_distributions_full} shows the rank
distributions of metaheuristics on all instances. All unsuccessful
metaheuristic-representation pairs but one use direct representation.
Only SA has been unable to succeed with either representation. For all
metaheuristics but GA, linear representation dominates direct
representation. Tab.~\ref{tab:fixed_target_rank_distributions_low}
shows the rank distributions of metaheuristics on low dimension
instances. Only SA with direct representation has been unable to
succeed. For all metaheuristics but GA, PBIL, UMDA, and MIMIC, linear
representation dominates direct representation.
Tab.~\ref{tab:fixed_target_rank_distributions_medium} shows the rank
distributions of metaheuristics on medium instances. The unsuccessful
metaheuristics are almost the same as for the full set of instances.
The difference with the full set of instances is that, in this
experiment, MIMIC has been successful. For all metaheuristics but GA
and MIMIC, linear representation dominates direct representation.

\begin{table}
  \centering
  \begin{tabular}{@{} l >{{\nprounddigits{0}}}n{2}{0}>{{\nprounddigits{2}}}n{2}{2}>{{\nprounddigits{1}}}n{2}{1}>{{\nprounddigits{2}}}n{2}{2}>{{\nprounddigits{0}}}n{2}{0} >{{\nprounddigits{1} \npunit{\%}}}n{3}{1} @{}}
\toprule
{Algorithm} & \multicolumn{5}{l}{{Rank}} & {Success} \\
\midrule
& {min} & {$Q_1$} & {med.} & {$Q_3$} & {max} & \\
\midrule
rls-linear & 1.000000 & 1.000000 & 1.000000 & 2.000000 & 14.000000 & 100\\
ea-1p1-linear & 1.000000 & 2.000000 & 2.000000 & 3.000000 & 18.000000 & 100\\
rls-direct & 1.000000 & 3.000000 & 3.000000 & 4.000000 & 21.000000 & 100\\
p3-linear & 4.000000 & 5.000000 & 6.000000 & 8.000000 & 22.000000 & 100\\
hc-linear & 1.000000 & 5.000000 & 7.000000 & 8.000000 & 15.000000 & 100\\
p3-direct & 4.000000 & 6.000000 & 7.000000 & 8.000000 & 22.000000 & 100\\
ltga-linear & 3.000000 & 6.000000 & 10.000000 & 13.500000 & 22.000000 & 100\\
ga-direct & 4.000000 & 9.000000 & 10.000000 & 11.000000 & 21.000000 & 100\\
ga-linear & 3.000000 & 10.000000 & 11.000000 & 13.000000 & 20.000000 & 100\\
ea-10p1-linear & 10.000000 & 12.000000 & 14.000000 & 15.250000 & 22.000000 & 100\\
ea-10p1-direct & 3.000000 & 13.000000 & 14.000000 & 16.000000 & 20.000000 & 100\\
umda-linear & 10.000000 & 12.000000 & 15.500000 & 18.250000 & 20.000000 & 100\\
ltga-direct & 3.000000 & 14.000000 & 16.000000 & 17.250000 & 21.000000 & 100\\
mimic-linear & 7.000000 & 14.000000 & 17.000000 & 20.000000 & 22.000000 & 100\\
pbil-linear & 8.000000 & 15.750000 & 17.500000 & 21.000000 & 22.000000 & 100\\
mimic-direct & 4.000000 & 7.000000 & 9.000000 & 10.750000 & 21.000000 & 99\\
sa-linear & 2.000000 & 12.250000 & 18.500000 & 20.000000 & 22.000000 & 98\\
hc-direct & 2.000000 & 13.750000 & 17.000000 & 19.250000 & 22.000000 & 98\\
pbil-direct & 5.000000 & 11.000000 & 12.500000 & 18.000000 & 21.000000 & 98\\
sa-direct & 1.000000 & 18.000000 & 21.000000 & 22.000000 & 22.000000 & 97\\
umda-direct & 5.000000 & 7.750000 & 13.500000 & 20.000000 & 22.000000 & 97\\
ea-1p1-direct & 2.000000 & 3.000000 & 4.000000 & 20.000000 & 22.000000 & 96\\
\bottomrule
\end{tabular}

  \caption{Rank statistics (runtime based) of metaheuristics on all
    instances.}
  \label{tab:fixed_target_rank_distributions_full}
\end{table}

\begin{table}
  \centering
  \begin{tabular}{@{} l >{{\nprounddigits{0}}}n{2}{0}>{{\nprounddigits{2}}}n{2}{2}>{{\nprounddigits{1}}}n{2}{1}>{{\nprounddigits{2}}}n{2}{2}>{{\nprounddigits{0}}}n{2}{0} >{{\nprounddigits{1} \npunit{\%}}}n{3}{1} @{}}
\toprule
{Algorithm} & \multicolumn{5}{l}{{Rank}} & {Success} \\
\midrule
& {min} & {$Q_1$} & {med.} & {$Q_3$} & {max} & \\
\midrule
rls-linear & 1.000000 & 1.000000 & 1.000000 & 2.500000 & 19.000000 & 100\\
ea-1p1-linear & 1.000000 & 2.000000 & 2.000000 & 3.250000 & 16.000000 & 100\\
rls-direct & 2.000000 & 3.000000 & 3.000000 & 4.000000 & 17.000000 & 100\\
ea-1p1-direct & 1.000000 & 3.000000 & 4.000000 & 4.000000 & 21.000000 & 100\\
hc-linear & 1.000000 & 5.000000 & 5.500000 & 7.000000 & 12.000000 & 100\\
p3-linear & 5.000000 & 5.750000 & 6.500000 & 14.500000 & 22.000000 & 100\\
p3-direct & 6.000000 & 7.000000 & 8.000000 & 10.000000 & 22.000000 & 100\\
umda-direct & 4.000000 & 7.000000 & 8.000000 & 12.000000 & 19.000000 & 100\\
mimic-direct & 9.000000 & 9.000000 & 11.000000 & 14.250000 & 20.000000 & 100\\
ga-direct & 9.000000 & 10.000000 & 11.000000 & 14.500000 & 21.000000 & 100\\
sa-linear & 2.000000 & 6.000000 & 11.500000 & 18.000000 & 22.000000 & 100\\
pbil-direct & 3.000000 & 11.000000 & 12.000000 & 12.250000 & 15.000000 & 100\\
ga-linear & 7.000000 & 11.000000 & 12.000000 & 13.000000 & 19.000000 & 100\\
hc-direct & 2.000000 & 8.750000 & 13.000000 & 15.500000 & 22.000000 & 100\\
ea-10p1-linear & 2.000000 & 13.000000 & 14.000000 & 15.250000 & 21.000000 & 100\\
ea-10p1-direct & 3.000000 & 14.000000 & 16.000000 & 17.000000 & 21.000000 & 100\\
ltga-linear & 7.000000 & 12.750000 & 16.500000 & 18.000000 & 20.000000 & 100\\
ltga-direct & 1.000000 & 15.250000 & 17.000000 & 18.000000 & 22.000000 & 100\\
umda-linear & 8.000000 & 15.250000 & 18.000000 & 20.000000 & 21.000000 & 100\\
mimic-linear & 6.000000 & 17.250000 & 20.000000 & 21.000000 & 22.000000 & 100\\
pbil-linear & 1.000000 & 14.250000 & 20.500000 & 21.000000 & 22.000000 & 100\\
sa-direct & 5.000000 & 15.750000 & 19.000000 & 22.000000 & 22.000000 & 99\\
\bottomrule
\end{tabular}

  \caption{Rank statistics (runtime based) of metaheuristics on low
    dimension instances.}
  \label{tab:fixed_target_rank_distributions_low}
\end{table}

\begin{table}
  \centering
  \begin{tabular}{@{} l >{{\nprounddigits{0}}}n{2}{0}>{{\nprounddigits{2}}}n{2}{2}>{{\nprounddigits{1}}}n{2}{1}>{{\nprounddigits{2}}}n{2}{2}>{{\nprounddigits{0}}}n{2}{0} >{{\nprounddigits{1} \npunit{\%}}}n{3}{1} @{}}
\toprule
{Algorithm} & \multicolumn{5}{l}{{Rank}} & {Success} \\
\midrule
& {min} & {$Q_1$} & {med.} & {$Q_3$} & {max} & \\
\midrule
rls-linear & 1.000000 & 1.000000 & 1.000000 & 1.000000 & 2.000000 & 100\\
ea-1p1-linear & 1.000000 & 2.000000 & 2.000000 & 2.000000 & 3.000000 & 100\\
rls-direct & 2.000000 & 3.000000 & 3.000000 & 4.000000 & 9.000000 & 100\\
p3-linear & 4.000000 & 5.000000 & 5.500000 & 6.000000 & 10.000000 & 100\\
ltga-linear & 3.000000 & 5.000000 & 6.000000 & 7.250000 & 12.000000 & 100\\
p3-direct & 4.000000 & 5.000000 & 6.500000 & 8.000000 & 8.000000 & 100\\
hc-linear & 3.000000 & 6.000000 & 8.000000 & 9.250000 & 12.000000 & 100\\
mimic-direct & 4.000000 & 7.000000 & 8.000000 & 9.000000 & 11.000000 & 100\\
ga-direct & 6.000000 & 9.000000 & 9.000000 & 10.000000 & 12.000000 & 100\\
ga-linear & 8.000000 & 10.000000 & 10.000000 & 13.000000 & 13.000000 & 100\\
ea-10p1-linear & 11.000000 & 11.000000 & 12.000000 & 14.250000 & 15.000000 & 100\\
umda-linear & 10.000000 & 11.750000 & 13.000000 & 15.250000 & 17.000000 & 100\\
ea-10p1-direct & 11.000000 & 13.000000 & 13.000000 & 14.000000 & 16.000000 & 100\\
mimic-linear & 14.000000 & 14.000000 & 15.000000 & 17.000000 & 19.000000 & 100\\
ltga-direct & 14.000000 & 15.000000 & 16.000000 & 16.000000 & 19.000000 & 100\\
pbil-linear & 15.000000 & 15.000000 & 16.000000 & 18.000000 & 20.000000 & 100\\
sa-linear & 17.000000 & 18.000000 & 19.000000 & 20.250000 & 22.000000 & 98\\
hc-direct & 17.000000 & 17.000000 & 19.500000 & 21.000000 & 22.000000 & 96\\
sa-direct & 18.000000 & 18.750000 & 21.000000 & 21.250000 & 22.000000 & 96\\
pbil-direct & 9.000000 & 12.000000 & 20.000000 & 20.000000 & 22.000000 & 96\\
umda-direct & 4.000000 & 6.750000 & 19.000000 & 21.000000 & 21.000000 & 95\\
ea-1p1-direct & 3.000000 & 4.000000 & 18.500000 & 22.000000 & 22.000000 & 92\\
\bottomrule
\end{tabular}

  \caption{Rank statistics (runtime based) of metaheuristics on medium
    dimension instances.}
  \label{tab:fixed_target_rank_distributions_medium}
\end{table}

\section{Conclusion}
\label{sec:conclusion}

We have proposed a linear representation for categorical values in
binary domains. It mostly targets evolutionary algorithms and other
metaheuristics expressed in terms of binary domains. It preserves the
neighborhood relations between categorical values. Every value can be
reached with a single mutation. This requirement has, in turn, lead to
an unexpected connexion with coding theory.

Linear representation has been paired with 11 standard metaheuristics
and applied to Sudoku puzzles. In fixed-budget experiments and
empirical cumulative distribution functions, high dimension instances
have been used to rank metaheuristic-representation pairs according to
the quality of their solutions. Linear representation has shown a
clear advantage over direct representation with all metaheuristics but
MIMIC, LTGA, and P3. Only in the case of P3 has direct representation
overtaken linear representation by a significant margin within the
considered budget. In fixed-target experiments, small dimension
instances have been used to rank metaheuristic-representation pairs
according to runtime. Linear representation has surpassed direct
representation with all metaheuristics but GA and MIMIC.

One drawback of linear representation is its size, which is linear in
the number of categorical values but exponential in the size of direct
representation. This could explain some of the negative experimental
results as an increased size of the search space usually implies a
degraded performance in terms of runtime or quality of solutions.

The results in this paper have to be confirmed in the context of other
problems, preferably real-world ones. The influence of the number of
categories and the number of categorical variables on the performance
of metaheuristic-representation pairs are of particular interest.

\bibliographystyle{plain}

\bibliography{bibliography}

\end{document}